\documentclass[11pt,a4paper]{article}
\usepackage[hyperref]{emnlp2018}
\usepackage{times}
\usepackage{latexsym}

\usepackage{url}
\usepackage{graphicx}
\usepackage{caption}
\usepackage{subcaption}
\usepackage{float}
\usepackage{amsfonts}
\usepackage{amsmath}
\usepackage{multirow}
\usepackage{threeparttable}
\DeclareMathOperator*{\argmax}{arg\,max}
\aclfinalcopy 

\title{Improving Word Representations: \\A Sub-sampled Unigram Distribution for Negative Sampling}

\author{Wenxiang Jiao$^\dagger$, Irwin King$^\dagger$, {\rm and} Michael R. Lyu$^\dagger$ \\
  $^\dagger$~Department of Computer Science and Engineering, \\
  The Chinese University of Hong Kong, HKSAR, China \\
  {\tt \{wxjiao,king,Lyu\}@cse.cuhk.edu.hk} \\}

\date{}

\begin{document}
\maketitle
\begin{abstract}
  Word2Vec is the most popular model for word representation and has been widely investigated in literature. However, its noise distribution for negative sampling is decided by empirical trials and the optimality has always been ignored. We suggest that the distribution is a sub-optimal choice, and propose to use a sub-sampled unigram distribution for better negative sampling. Our contributions include: (1) proposing the concept of semantics quantification and deriving a suitable sub-sampling rate for the proposed distribution adaptive to different training corpora; (2) demonstrating the advantages of our approach in both negative sampling and noise contrastive estimation by extensive evaluation tasks; and (3) proposing a semantics weighted model for the MSR sentence completion task, resulting in considerable improvements. Our work not only improves the quality of word vectors but also benefits current understanding of Word2Vec.
\end{abstract}

\section{Introduction}
\label{sec:introduction}

The recent decade has witnessed the great success achieved by word representation in natural language processing (NLP). It proves to be an integral part of most other NLP tasks, in which words have to be vectorized before input to the models. High quality word vectors have boosted the performance of many tasks, such as named entity recognition~\citep{pennington2014glove,sienvcnik2015adapting}, sentence completion~\citep{DBLP:journals/corr/YogatamaFDS14,liu2015learning}, part-of-speech tagging~\citep{ling2015two,ling2015not}, sentiment analysis~\citep{tsvetkov2015evaluation,yu2017refining}, and machine translation~\citep{sutskever2014sequence,johnson2016google}. In a conventional way, word vectors are obtained from word-context co-occurrence matrices by either cascading the row and column vectors~\citep{lund1996producing} or applying singular value decomposition (SVD)~\citep{deerwester1990indexing}. However, these approaches are limited by their sub-optimal linear structure of vector space and the highly increased memory requirement when confronting huge vocabularies. Both problems have been solved by a popular model called Word2Vec~\citep{mikolov2013distributed}, which utilizes two shallow neural networks, {\em i.e.}, skip-gram and continuous bag-of-words, to learn word vectors from large corpora. The model is also capable of capturing interesting linear relationships between word vectors.

While Word2Vec makes a breakthrough in word representation, it has not been fully understood and its theoretical exploitation is still in demand. One aspect, which has always been ignored, is the choice of noise distribution for negative sampling. Word2Vec employs a smoothed unigram distribution with a power rate of 3/4 as the noise distribution. The decision is made by empirical trials but has been widely adopted in subsequent work~\citep{levy2015improving,ling2015two,yang2017simple,bamler2017dynamic}. However, the quality of learned word vectors is sensitive to the choice of noise distribution~\citep{gutmann2010noise,levy2015improving} when using a moderate number (5 to 15) of negative samples, which is a common strategy for the tradeoff between vector quality and computation costs.   

In this paper, we propose to employ a sub-sampled unigram distribution for negative sampling and demonstrate its capability of improving the linear relationships between word vectors. Our contributions include three aspects:
(1) We propose the concept of semantics quantification and derive a suitable sub-sampling rate for the proposed distribution.
(2) We demonstrate the advantages of our noise distribution in both negative sampling and noise contrastive estimation by extensive experiments.
(3) We propose a semantics weighted model for the MSR sentence completion task, resulting in considerable improvements.

\section{Word2Vec}
\label{sec:Word2Vec}

\subsection{Architectures}
\label{ssec: Architectures}

Firstly, we briefly introduce the two architectures, {\em i.e.}, skip-gram (SG) and continuous bag-of-words (CBOW) in Word2Vec~\citep{mikolov2013distributed}.
For a corpus with a word sequence $w_{1}, w_{2}, \cdots, w_{T}$, skip-gram predicts the context word $w_{t+j}$ given the center word $w_t$, and maximizes the average log probability, 
\begin{equation}
\frac{1}{T}\sum_{t=1}^{T}\sum_{-c \leq j \leq c, j \neq 0}\log p(w_{t+j}|w_{t}) \label{SG_obj},
\end{equation}
where $c$ is the size of context window, and $p(w_{t+j}|w_{t})$ is defined by the full softmax function,
\begin{equation}
p(w_{O}|w_{I}) = \frac{\exp\left(v_{w_{O}}'^{\top}v_{w_{I}}\right)}
{\sum_{w=1}^{|V|}\exp\left(v_{w}'^{\top}v_{w_{I}}\right)}  \label{softmax},
\end{equation}
\noindent
where $v_{w}$ and $v_{w}'$ are the vectors of the ``input'' and ``output'' words, and $|V|$ is the size of vocabulary.

As for CBOW, it predicts the center word based on the context words. The input vector is usually the average of the context words' vectors, {\em i.e.}, $v_{w_{I}} = \frac{1}{2c} \sum_{-c \leq j \leq c, j \neq 0} v_{w_{t+j}}$.

\begin{figure}[t]
\centering
\captionsetup{belowskip=-1em}
\includegraphics[width=0.45\textwidth]{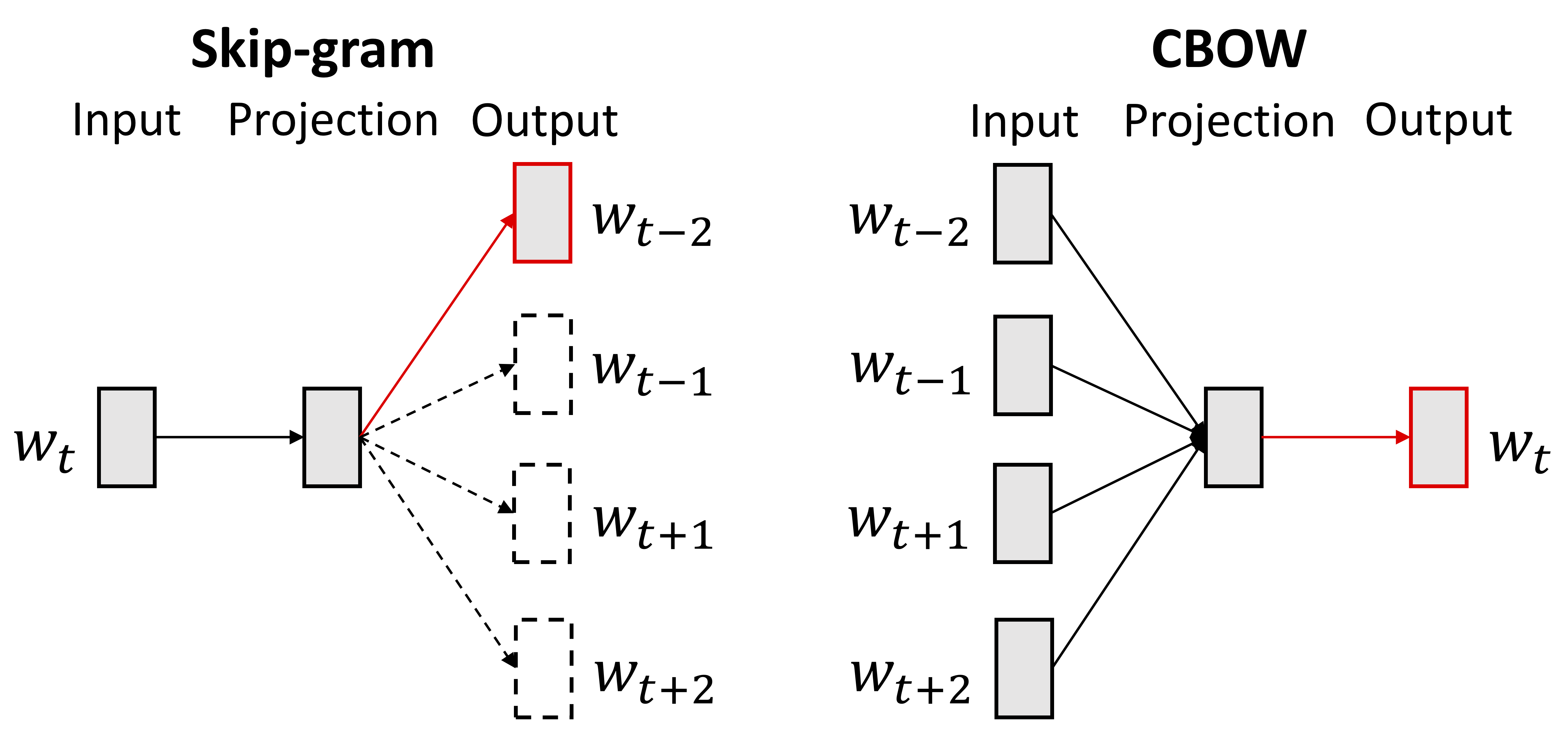}
\caption{ Illustration of the skip-gram and continuous bag-of-words (CBOW) architectures.} 
\label{Fig_arch}
\end{figure}

\subsection{Negative Sampling}
\label{ssec: Negative sampling}

For large vocabularies, it is inefficient to compute the full softmax function in Eq.\,(\ref{softmax}). To tackle this problem, Word2Vec utilizes negative sampling to distinguish the real output word from $k$ noise words,
\begin{equation}
\log \sigma(v_{w_{O}'^{\top}} v_{w_{I}}) + \sum_{i=1}^{k} \mathbb{E}_{ P_{n}(w_{i})}\left[\log\sigma(-v_{w_{i}}'^{\top}v_{w_{I}})\right]  \label{(neg)},
\end{equation}
\noindent where $\sigma(x) = \frac{1}{1 + \exp(-x)}$, and $P_n$ is the so-called noise distribution, representing the probability for a word to be sampled as a noise word. The smoothed unigram distribution used in Word2Vec is expressed as,
\begin{equation}
P_n(w_{i})= \frac{f(w_{i})^{3/4}}{\sum_{j=1}^{|V|}f(w_{j})^{3/4}}, \label{Pn}
\end{equation}
where $f(w_i)$ is the frequency of word $w_i$.
\subsection{Sub-sampling}

Sub-sampling is a process in Word2Vec for randomly deleting the most frequent words during training, since they are usually stop words with less information than infrequent ones. During sub-sampling, the probability that a word $w_i$ should be kept is defined as,
\begin{equation}
P_{keep} (w_{i}) = \left(\sqrt[]{\frac{\hat f(w_{i})}{t}}+1\right)\frac{t}{\hat f(w_{i})}  \label{Psub},
\end{equation}
where $\hat f(w_i)$ is the normalized word frequency of $w_i$, and $t$ is called the sub-sampling rate typically between $10^{-5}$ and $10^{-3}$. The process does not delete infrequent words.

\section{Related Work}
\label{ssec: Related Work}

{\bf Unigram.} A noise distribution is recommended to be close to the distribution of the real data in noise contrastive estimation (NCE)~\citep{gutmann2010noise}. Such guidance finds its earliest application for training language models by \citet{mnih2012fast}, demonstrating that the unigram distribution works better than a uniform distribution. This choice is also adopted in some other work~\citep{mnih2013learning,vaswani2013decoding,xiao2013domain,baltescu2014pragmatic}. However, the performance of models is limited due to the inadequate training of infrequent words~\citep{chen2015strategies,labeau2017experimental}.

{\bf Smoothed Unigram.} The smoothed unigram distribution in Word2Vec~\citep{mikolov2013distributed} solves this problem because it gives more chances for infrequent words to be sampled. However, the required power rate is decided empirically, and may need adjustment for different scenarios~\citep{bojanowski2016enriching,ai2016analysis}. \citet{labeau2017experimental} even propose to use a bigram distribution after studying the power rate, but it is infeasible for large corpora. Besides, the smoothed unigram distribution also changes the lexical structure of infrequent words, which could be a reason for the limited quality of word vectors.

\section{Sub-sampled Unigram Distribution}
\label{sec: Sub-sampled}

We believe a sub-sampled unigram distribution is better for negative sampling since it reduces the amount of frequent words and also maintains the lexical structure of infrequent words. To our best knowledge, we are the first to employ such a noise distribution for negative sampling. Beyond this, we propose a approach to derive the sub-sampling rate that is adaptive to different corpora (Table\,\ref{Tcorpora}).

\subsection{Critical Word}
\label{sec:Critical}

We start our analysis by recalling the probability in Eq.\,(\ref{Psub}) of a word to be kept during sub-sampling. Obviously, we need to choose the sub-sampling rate $t$ to decide the noise distribution. Although empirically selecting a sub-sampling rate can result in improvements (Table\,\ref{Tnce}), we aim to derive the sub-sampling rate adaptive to different corpora. To accomplish this, we firstly introduce a concept {\em critical word} denoted by $w_{crt}$, which is the word with $P_{keep}(w_{crt})=1$. The critical word indicates that words with frequencies lower than it will not be deleted during sub-sampling. It is uniquely decided by the sub-sampling rate. Thus, if we select the critical word with certain properties at first, we are able to obtain a suitable sub-sampling rate in return.

The basic rule for us to select the critical word is to find a word with balanced semantic and syntactic information. We prefer not to delete words with relatively more semantic information. Now, the problem is how to measure these two kinds of information a word possesses.

\begin{figure}[t]
\centering
\captionsetup{belowskip=-1em}
\includegraphics[width=0.44\textwidth]{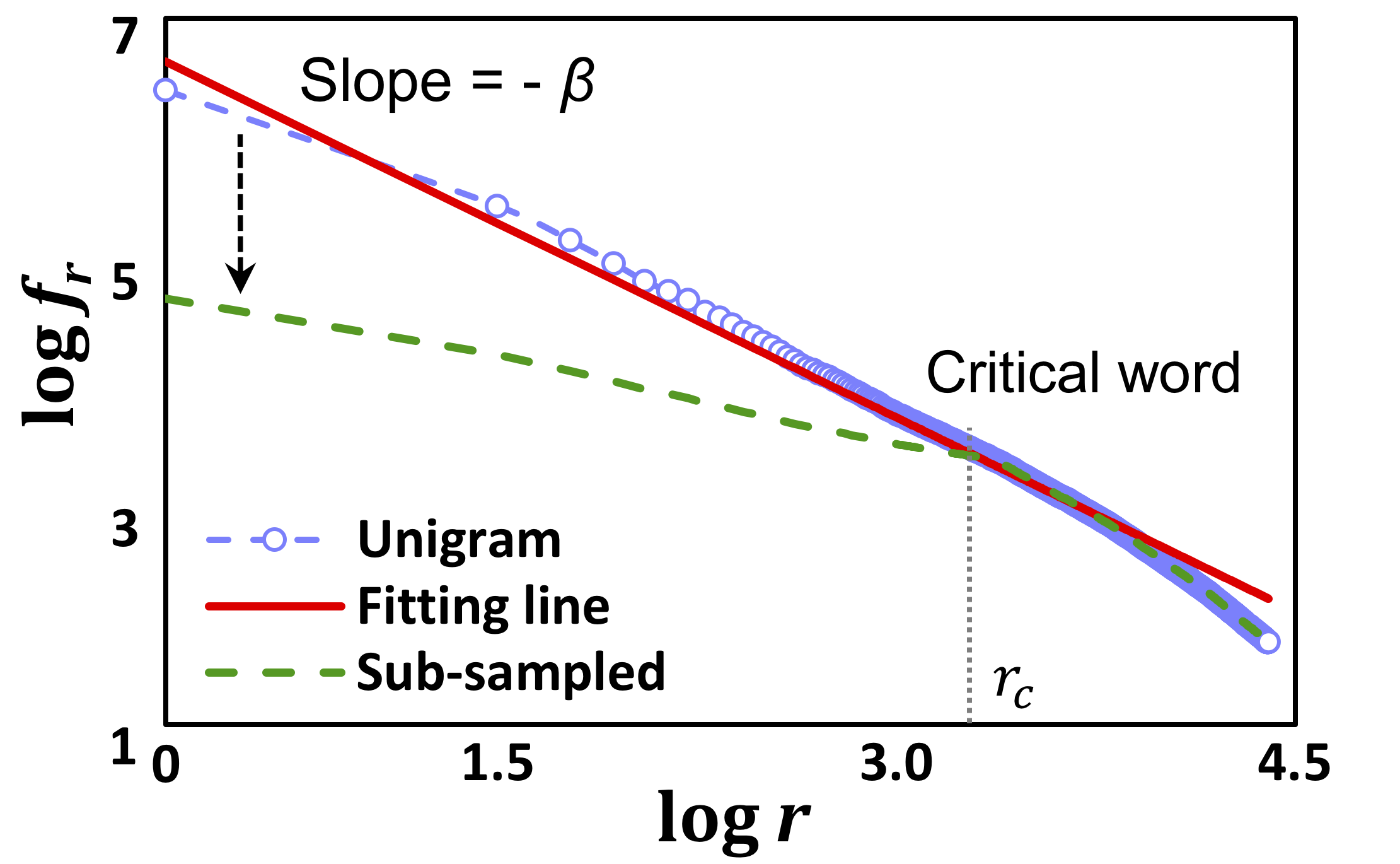}
\caption{Illustration of a unigram distribution, the fitting line, and the sub-sampled version.}
\end{figure}

\subsection{Semantics Quantification}
\label{ssec: SemQ}

In order to quantify the semantic and syntactic information of words, we consider two observations: (1) frequent words are more likely to be function words with more syntactic information; (2) infrequent words are more likely to be content words with more semantic information~\citep{hochmann2010word}.
Thus, for the $r$-th most frequent word $w$, the quantity of its semantic and syntactic information $I_{sem}^{w}$ and $I_{syn}^{w}$, can be described as,
\begin{equation}
I_{sem}^{w} = F_1(r) ,\ I_{syn}^{w} = F_2(f_r) ,
\end{equation}
where $F_1(r)$ and $F_2(f_r)$ are monotonically increasing functions of the ranking $r$ and the frequency $f_r$, respectively. One can tell that the functions capture the properties of the observations.

On the other hand, we require that the total quantity of semantic and syntactic information, denoted by $I_{tot}^{w}$ is fixed for all words, {\em i.e.},
\begin{equation}
I_{tot}^{w}= F_1(r) + F_2(f_r) = \mathrm{const}_1,
\label{Itot}
\end{equation}
where $\mathrm{const}_1$ is a constant. We rewrite Eq.\,(\ref{Itot}) into an exponential form as the following,
\begin{equation}
\exp(F_2(f_r)) = \frac{\exp(\mathrm{const}_1)}{\exp(F_1(r))}.
\label{Irewrite}
\end{equation}
This expression leads us to a well known power law called Zipf's law~\citep{Zipf1950-ZIPHBA}, which approximates the relationship between $f_r$ and $r$ as,
\begin{equation}
f_r = \frac{\gamma}{r^\beta}, \label{zipf1}
\end{equation}
\noindent where $\gamma, \beta$ are constants and $\beta \approx 1$.
Consequently, we can decide the form of the functions $F_1(r)$ and $F_2(f_r)$ as, 
\begin{equation}
F_1(r) = \log r ,\ F_2(f_r) = \log f_r .
\end{equation}
Obviously, the $\log$ form functions satisfy the definition we made before. As a results, the total information becomes $\log\gamma$ given $\beta \approx 1$.

\subsection{Expression of Sub-sampling Rate}
\label{ssec: ExpSub}

Now, given the quantified information, we are able to decide the critical word satisfying the condition
\begin{equation}
I_{sem}^{w_{r_c}} = I_{syn}^{w_{r_c}} . \label{cond_fc}
\end{equation}

Combined with Eq.\,(\ref{zipf1}), we obtain the frequency of the critical word 
\begin{equation}
\log f_{r_c} = \frac{\log \gamma}{1 + \beta}, \label{fc}
\end{equation}
where $r_c$ is the ranking of the critical word. Meanwhile, we know the probability of the critical word $w_{crt}$ to be kept should be exactly $P_{keep}^{t_c} (w_{crt})=1$. Thus, with Eq.\,(\ref{Psub}) and Eq.\,(\ref{fc}),  the sub-sampling rate for our noise distribution is expressed as
\begin{equation}
t_c = \frac{4 \hat f_{r_c}}{(1 + \sqrt[]{5})^2}. \label{tc}
\end{equation}
Note that we use $t_c$ to distinguish from the sub-sampling rate $t$ applied for the training corpus.

\subsection{Constants Estimation}
\label{ssec: ConstEst}

As for the estimation of constants $\gamma$ and $\beta$, we provide two choices:
\\[1ex]
(1) \textbf{wLSE-1}. We use weighted least squares estimation (wLSE) to estimate the two constants. Since more data are located at higher positions in $\log r$ axis, wLSE with a weight of $\frac{1}{r}$ for the {\em r}-th most frequent word makes sure the trend of line can be well fit. The estimated constants are 
\begin{align}
\hat \beta = - & \frac{\left\langle\log r \log f_r\right\rangle - \left\langle\log r\right\rangle \left\langle \log f_r\right\rangle}{\left\langle(\log r)^2\right\rangle - \left(\left\langle\log r\right\rangle\right)^2}, \\
& \log\hat\gamma = \left\langle \log f_r\right\rangle + \hat\beta \left\langle \log r\right\rangle, \label{hat_b1}
\end{align}
where $\left\langle x \right\rangle$ denotes the weighted average of $x$ such that $\left\langle x \right\rangle = \sum_{r=1}^{|V|}\frac{x}{r} / \sum_{r=1}^{|V|}\frac{1}{r}$. 
\\[1ex]
(2) \textbf{wLSE-2.} We use wLSE with a condition that the fitting line passes through the point $(\log 1, \log f_1)$. This method engages the most frequent word to further control the trend of the line. As a result, $\hat\gamma = f_1$ and 
\begin{equation}
\hat\beta = - \frac{\left\langle\log r \left(\log f_r -\log f_1\right)\right\rangle}{\left\langle(\log r)^2\right\rangle}. \label{hat_a2}
\end{equation}

Now, we can write down the expression of the sub-sampled unigram distribution 
\begin{equation}
P_n^{sub}(w_i) = \frac{\alpha_i f(w_i)}{\sum_{j=1}^{|V|}\alpha_i f(w_j)},
\end{equation}
where $\alpha_i$ satisfies
\begin{eqnarray}
\alpha_i = \left\{
\begin{array}{ccr}
P_{keep}^{t_c}(w_i) &\mathrm{if}\ P_{keep}^{t_c}(w_i) < 1 \\
1 &\mathrm{otherwise}
\end{array}\right..
\end{eqnarray}
Note that we use $P_n^{sub}$ to distinguish from the original noise distribution $P_n$ in Word2Vec.

\subsection{Discussions}
\label{ssec: Discussions}

In semantics quantification, the modeling of word distribution is not limited to zipf's law. We adopt it because of its popularity and conciseness. There could be other choices~\citep{mandelbrot1953informational,piantadosi2014zipf}, and the expression of $t_c$ needs modification accordingly. Besides, one can either use the chosen law to decide the critical word or just search through the unigram distribution to find it.

\section{Experiments}
\label{sec: Experiments}

To show the advantages of our noise distribution, we conduct experiments on three evaluation tasks. While the word analogy task~\citep{mikolov2013distributed} is our focus for testing the linear relationships between word vectors, we also evaluate the learned word vectors on the word similarity task~\citep{pennington2014glove} and the synonym selection task~\citep{liu2015learning}.

In the following, we firstly describe the experimental setup including baselines, training corpora and details. Next, we report experimental results for the three NLP tasks. At last, we introduce the semantics weighted model proposed for the MSR sentence completion task~\citep{zweig2012computational}.

\subsection{Experimental Setup}
\label{ssec: ExpSetup}

\subsubsection{Baselines}
\label{sssec: Baselines}

We train the two models, SG and CBOW, using the original noise distribution and other two obtained by our approach, specifically,
\\[1ex]
(1) {\em Uni$^{3/4}$.} The smoothed unigram distribution proposed by~\citet{mikolov2013distributed}.
\\[1ex]
(2) {\em Sub$^{L1}$.} The sub-sampled uinigram distribution, of which the threshold $t_c$ is estimated by wLSE-1.
\\[1ex]
(3) {\em Sub$^{L2}$.} The sub-sampled uinigram distribution, of which the threshold $t_c$ is estimated by wLSE-2.

\subsubsection{Training Corpora}
\label{sssec: Corpora}

Our training corpora come from four sources, described as below:
\\[1ex]
(1) \textbf{BWLM.} The ``One Billion Word Language Modeling Benchmark''\footnote{\url{http://www.statmt.org/lm-benchmark}}, which is already pre-processed and has almost 1 billion tokens.
\\[1ex]
(2) \textbf{Wiki10.} The April 2010 snapshot of the Wikipedia corpus\footnote{\url{http://www.psych.ualberta.ca/~westburylab/downloads/westburylab.wikicorp.download.html}} with a total of about 2 million articles and 1 billion tokens.
\\[1ex]
(3) \textbf{UMBC.} The UMBC WebBase corpus\footnote{\url{http://ebiquity.umbc.edu/blogger/2013/05/01/umbc-webbase-corpus-of-3b-english-words}} from the Stanford WebBase project’s February 2007 Web crawl, with over 3 billion tokens.
\\[1ex]
(4) \textbf{MSR.} The MSR corpus containing 5 Conan Doyle Sherlock Holmes novels\footnote{\url{https://www.microsoft.com/en-us/research/project/msr-sentence-completion-challenge}} with about 50 million tokens.

The first three large corpora are used for word similarity, synonym selection, and word analogy tasks. The MSR corpus is designated for the MSR sentence completion task. We pre-process the corpora by converting all words into lowercase and removing all the non-alphanumeric. The number of remaining tokens for each corpus is listed in the column {\bf Size} of Table\,\ref{Tcorpora}. Vocabularies are built by discarding words whose occurrences are less than the threshold shown in the column {\bf Mcn}. The column {\bf Vocab} represents the sizes of the resulted vocabularies. The rightmost two columns are the sub-sampling rates for our noise distribution by the wLSE-1 and wLSE-2 estimations, respectively. The values are $10^6$ times of the true ones for readability.

\begin{table}[t]
\centering
\captionsetup{belowskip=-1em}
\begin{tabular}{|l|c|c|c|c|c|}
\hline \bf Corpus & \bf Size & \bf Mcn & \bf Vocab & $\boldsymbol t_c$-1 & $\boldsymbol t_c$-2  \\ \hline
BWLM & 0.7\,B & 20 & 195\,k & 3.10 & 3.17 \\
Wiki10 & 1.0\,B & 50 & 249\,k & 2.76 & 2.80 \\
UMBC & 3.0\,B & 50 & 267\,k & 1.33 & 1.39 \\
MSR & 47\,M & 5 & 77\,k & 13.2 & 13.1 \\
\hline
\end{tabular}
\caption{\label{Tcorpora} Information of the training corpora.}
\end{table}

\subsubsection{Training details}
\label{sssec: Training}

We implement the training of word vectors with the {\tt word2vec} tool\footnote{\url{https://code.google.com/archive/p/word2vec}}, in which the part of noise distribution is modified to support several choices.
For SG and CBOW, we set the vector dimensionality to 100, and the size of the context window to 5. We choose 10 negative samples for each training sample in the models. The models are trained using the stochastic gradient decent (SGD) algorithm with a linear decaying learning rate with an initial value of 0.025 in SG and 0.05 in CBOW. We train the models on the three large corpora for 2 epochs, and for MSR's Holmes novels the value may vary. Results in this paper are shown in percentages and each of them is the average result of 4 repeated experiments, unless otherwise stated.

\subsection{Task 1: Word Similarity Task}
\label{ssec: Task1}

\subsubsection{Task Description}
\label{sssec: Task1_Desp}

The task computes the correlation between the word similarity scores by human judgment and the word distances in vector space. We use Pearson correlation coefficient $\rho_p$ as the metric, the higher of which the better the word vectors are. The expression of $\rho_p$ is
\begin{equation}
\rho_p = \frac{\mathrm{Cov}(\phi, \hat\phi)}{\sqrt{\mathrm{Var}(\phi) \mathrm{Var}(\hat\phi)}} ,
\label{pearson}
\end{equation}
\noindent where $\phi$ and $\hat\phi$ are random variables for the word similarity scores by human judgment and the cosine distances between word vectors, respectively. Benchmark datasets for this task include RG~\citep{rubenstein1965contextual}, MC~\citep{miller1991contextual}, WS~\citep{finkelstein2001placing}, MEN~\citep{bruni2012distributional}, and RW~\citep{luong2013better}.

\subsubsection{Results}
\label{sssec: Task1_results}

We implement the task on the mentioned 5 datasets and show the results in the column {\bf Word Similarity} of Table\,\ref{Tresults}. At the first glance, our noise distributions {\em Sub$^{L1}$} and {\em Sub$^{L2}$} perform slightly better than {\em Uni$^{3/4}$}. Significant improvements can be achieved on two small datasets RG and MC, because they are more sensitive to the vector quality. Another observation is that CBOW is more affected by {\em Sub$^{L1}$} and {\em Sub$^{L2}$} than SG, if comparing results on RG and MC with Wiki10 corpus. These results show that our noise distributions have the potential as high as or even higher than the smoothed unigram distribution in learning good word vectors.

\begin{table*}[t!]
\centering
\captionsetup{belowskip=-1em}
\begin{threeparttable}
\begin{tabular}{|c|c|p{0.9cm}||p{0.55cm}|p{0.55cm}|p{0.55cm}|c|p{0.55cm}||c|c|p{0.55cm}||p{0.55cm}|p{0.55cm}|p{0.55cm}|}
\hline
\multirow{2}{*}{\bf Size}
& \multirow{2}{*}{\bf Model} 
& \multirow{2}{*}{\bf Noise}
& \multicolumn{5}{c||}{\bf Word Similarity}
& \multicolumn{3}{c||}{\bf Synonym Selection}
& \multicolumn{3}{c|}{\bf Word Analogy}\\ 
\cline{4-14}
 & & & \bf RG & \bf MC & \bf WS & \bf MEN & \bf RW & \bf LEX & \bf Toefl & \bf Tot & \bf Sem & \bf Syn & \bf Tot \\
\hline
\multirow{6}{*}{0.7\,B} 
 & \multirow{3}{*}{\tt sg} 
 & {\em Uni$^{3/4}$} & 62.1 & 64.4 & 62.5 & 66.8 & \underline{43.3} & 66.1 & \bf74.6 & 67.9 & 61.4 & \bf57.4 & 59.2 \\
 & & {\em Sub$^{L1}$} & 62.9 & 66.0 & \bf63.1 & \underline{67.1} & \bf43.3 & 67.9 & 73.6 & 69.1 & 62.8 & 56.8 & 59.5 \\
 & & {\em Sub$^{L2}$} & \bf63.0 & \bf66.8 & 62.8 & \bf67.1 & 43.2 & \bf68.3 & 73.6 & \bf69.4 & \bf63.5 & 56.9 & \bf59.9 \\
\cline{2-14}
 & \multirow{3}{*}{\tt cbow} 
 & {\em Uni$^{3/4}$} & 64.3 & 66.4 & 60.4 & 66.1 &	44.0 & 66.4 & \bf79.6 & 69.2 & 53.4 & 58.7 & 56.4 \\
 & & {\em Sub$^{L1}$} & 64.6 & 67.1 & \bf61.0 & \underline{66.7} & \bf44.1 & 67.4 & 78.2 & 69.7 & 57.4 & 59.8 & 58.7 \\
 & & {\em Sub$^{L2}$} & \bf65.7 & \bf67.9 & 60.7 & \bf66.7 & 43.7 & \bf68.3 & 79.2 & \bf70.7 & \bf58.0 & \bf60.4 & \bf59.3 \\
\hline
\multirow{6}{*}{1.0\,B}  
 & \multirow{3}{*}{\tt sg} 
 & {\em Uni$^{3/4}$} & 77.2 & 81.4 & 68.2 & 70.1 &43.3 & \bf65.9 & 82.8 & 69.6 & 61.6 & \bf58.8 & 60.1 \\
 & & {\em Sub$^{L1}$} & 77.2 & \bf81.9 & \bf68.7 & \bf70.5 & \bf43.6 & 65.6 & \bf86.3 & \bf70.2 & \bf64.2 & 58.6 & \bf61.1 \\
 & & {\em Sub$^{L2}$} & \bf77.3 & 81.5 & 68.4 & 70.4 & 43.5 & 64.7 & 84.4 & 69.1 & 63.9 & 58.7 & \underline{61.1} \\
\cline{2-14}
 & \multirow{3}{*}{\tt cbow} 
 & {\em Uni$^{3/4}$} & 76.2 & 76.9 & 68.7 & 70.6 & 44.1 & 68.8 & 82.8 & 71.9 & 65.0 & 61.3 & 63.0 \\
 & & {\em Sub$^{L1}$} & \bf77.4 & \bf80.3 & \bf69.3 & 71.0 & \bf44.6 & 67.7 & \bf84.0 & 71.3 & 67.4 & 62.2 & 64.6 \\
 & & {\em Sub$^{L2}$} & 76.8 & 80.0 & 69.2 & \bf71.2 &	44.3 & \bf69.5 & 80.8 & \bf72.0 & \bf68.4 & \bf62.7 & \bf65.3 \\
\hline
\multirow{6}{*}{3.0\,B} 
 & \multirow{3}{*}{\tt sg} 
 & {\em Uni$^{3/4}$} & 69.7 & 77.6 & 67.6 & 69.5 & 46.7 & 72.6 & 84.6 & 75.1 & 46.4 & 63.2 & 55.7 \\
 & & {\em Sub$^{L1}$} & 69.9 & 78.7 & \underline{67.8} & \underline{70.2} & \bf47.3 & \bf72.9 & 84.6 & \bf75.3 & \bf51.4 & \bf64.1 & \bf58.4 \\
 & & {\em Sub$^{L2}$} & \bf70.5 & \bf79.2 & \bf67.8 & \bf70.2 & 47.1 & 72.4 & \bf85.9 & 75.2 & 51.2 & 63.8 & 58.2 \\
\cline{2-14}
 & \multirow{3}{*}{\tt cbow} 
 & {\em Uni$^{3/4}$} & 72.7 & 77.5 & 67.3 & 71.0 &	48.2 & 76.5 & \underline{87.8} & 78.8 & 44.9 & 64.6 & 55.9 \\
 & & {\em Sub$^{L1}$} & 74.2 & 78.2 & 67.8 & \underline{71.5} &	\bf48.5 & 77.2 & \bf87.8 & 79.4 & 50.6 & 66.6 & 59.5 \\
 & & {\em Sub$^{L2}$} & \bf74.4 & \bf78.7 & \bf68.0 & \bf71.5 & 48.4 & \bf78.0 & 87.1 & \bf79.9 & \bf50.7 & \bf66.8 & \bf59.6 \\
\hline
\end{tabular}
\end{threeparttable}
\caption{\label{Tresults} Results of evaluation tasks on the learned word vectors, {\em i.e.}, word similarity, synonym selection, and word analogy. The sub-sampling rate for the training corpora is $10^{-4}$.}
\end{table*}

\subsection{Task 2: Synonym Selection Task}
\label{ssec: Task2}

\subsubsection{Task Description}
\label{sssec: Task2_Desp}

This task attempts to select the semantically closest word, from the candidate answers, to the stem word. For example, given the stem word ``costly'' and the candidate answers ``expensive, beautiful, popular, complicated'', the most similar word should be ``expensive''. For each candidate answer, we compute the cosine similarity score between its word vector and that of the stem word. The candidate answer with the highest score is our final answer for a question. Here we use the TOEFL dataset~\citep{landauer1997solution} with 80 synonym questions and the LEX\footnote{We collect the questions from two ebooks {\em 501 Synonym and Antonym Questions} and {\em 1001 Vocabulary \& Spelling Questions} provided on the eLearning platform LearningExpress. \url{https://www.learningexpresshub.com}} dataset with 303 questions collected by ourselves.

\subsubsection{Results}
\label{sssec: Task2_results}

We report the results of this task in the {\bf Synonym Selection} column of Table\,\ref{Tresults}. For all the noise distributions, the results are not stable on TOEFL dataset since it is quite small. Still, {\em Sub$^{L1}$} and {\em Sub$^{L2}$} have comparable performance with {\em Uni$^{3/4}$}. In particular, {\em Sub$^{L1}$} makes considerable improvements with Wiki10 corpus.
As for LEX dataset, {\em Sub$^{L1}$} and {\em Sub$^{L2}$} outperform {\em Uni$^{3/4}$} in both SG and CBOW models with BWLM corpus. With the other two corpora, {\em Sub$^{L2}$} performs better than {\em Sub$^{L1}$} and {\em Uni$^{3/4}$} using CBOW model. But again, the SG model appears to be less boosted by {\em Sub$^{L1}$} and {\em Sub$^{L2}$} in terms of the corresponding results. Considering the unbalanced number of questions in these two datasets, we provide the total results on TOEFL+LEX and conclude that our noise distributions are better than {\em Uni$^{3/4}$}.

\subsection{Task 3: Word Analogy Task}
\label{ssec: Task3}

\begin{figure*}[t!]
\centering
\captionsetup{belowskip=-1em}
\centering
\begin{subfigure}{0.3\textwidth} 
\includegraphics[width=\textwidth]{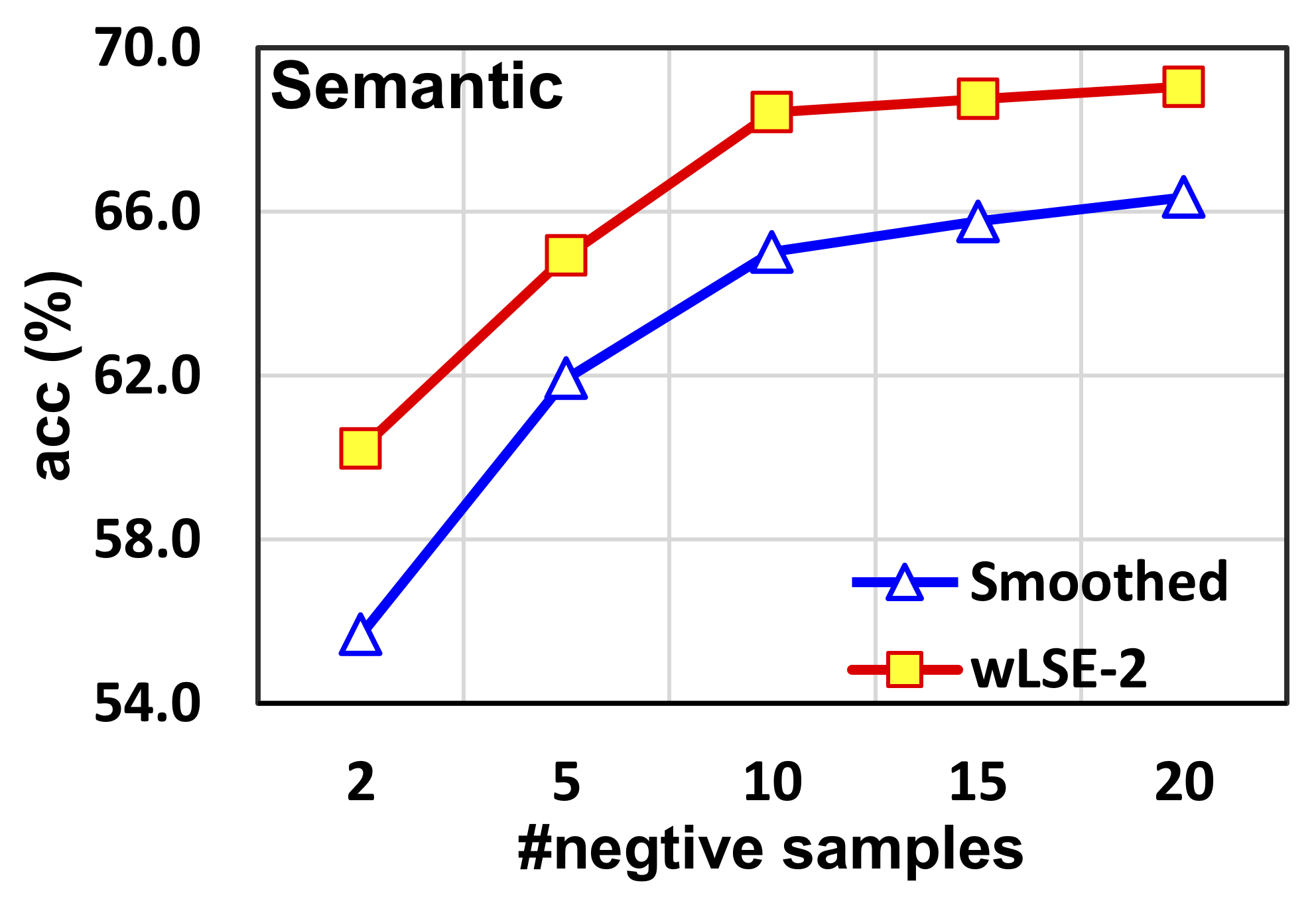} \\
\includegraphics[width=\textwidth]{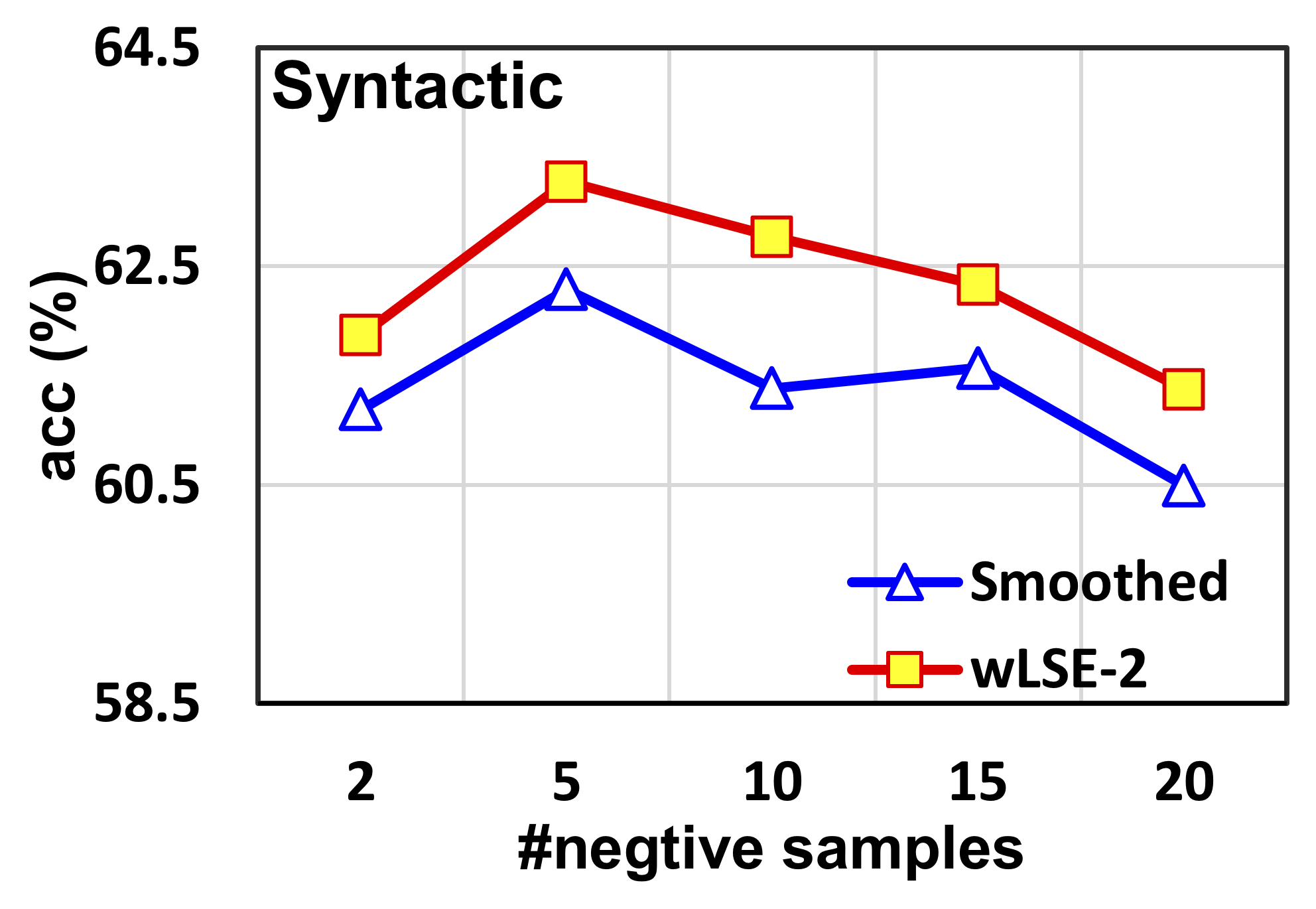}
\caption{CBOW}
\end{subfigure}
\hspace{0.6em}
\begin{subfigure}{0.3\textwidth} 
\includegraphics[width=\textwidth]{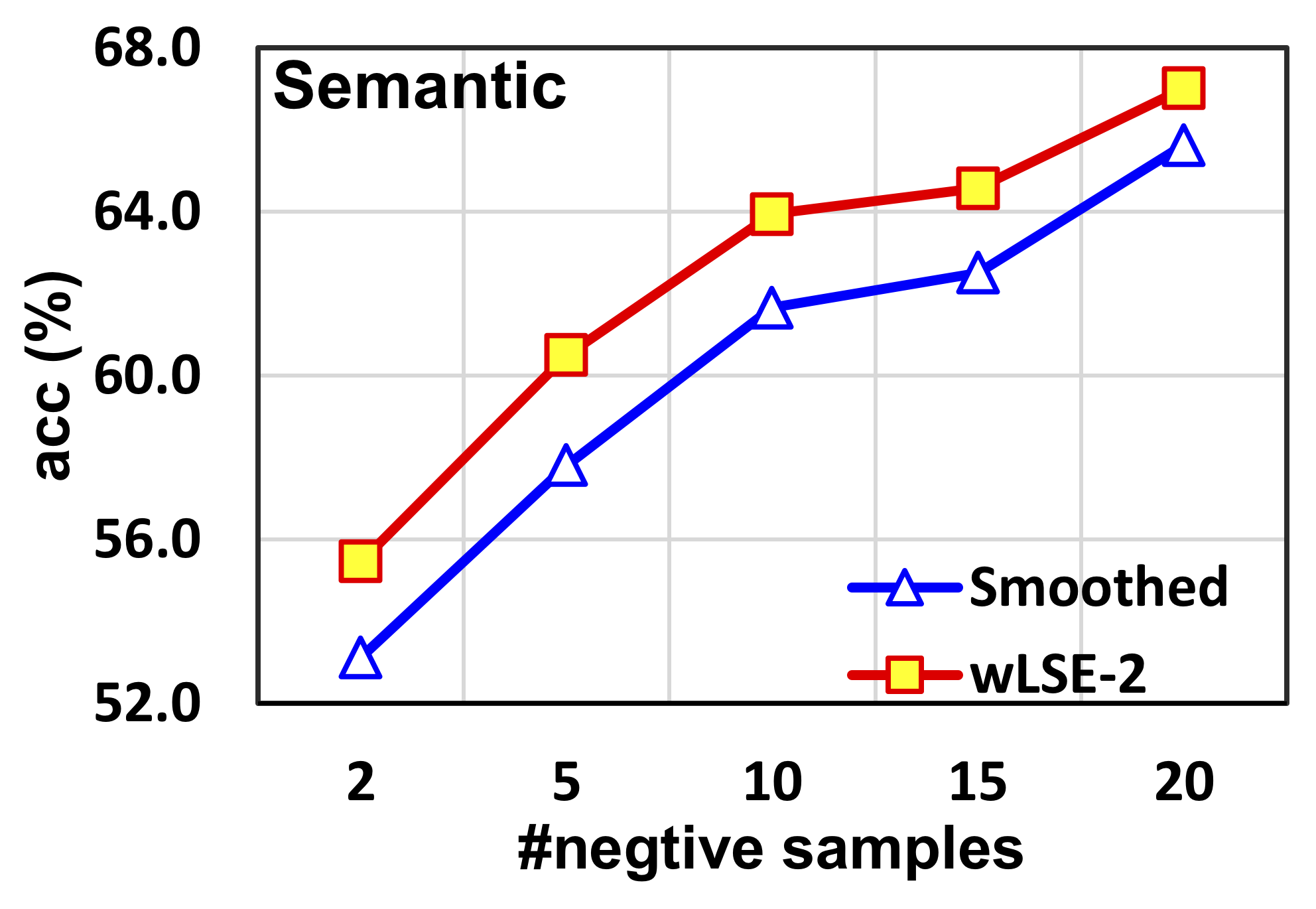} \\
\includegraphics[width=\textwidth]{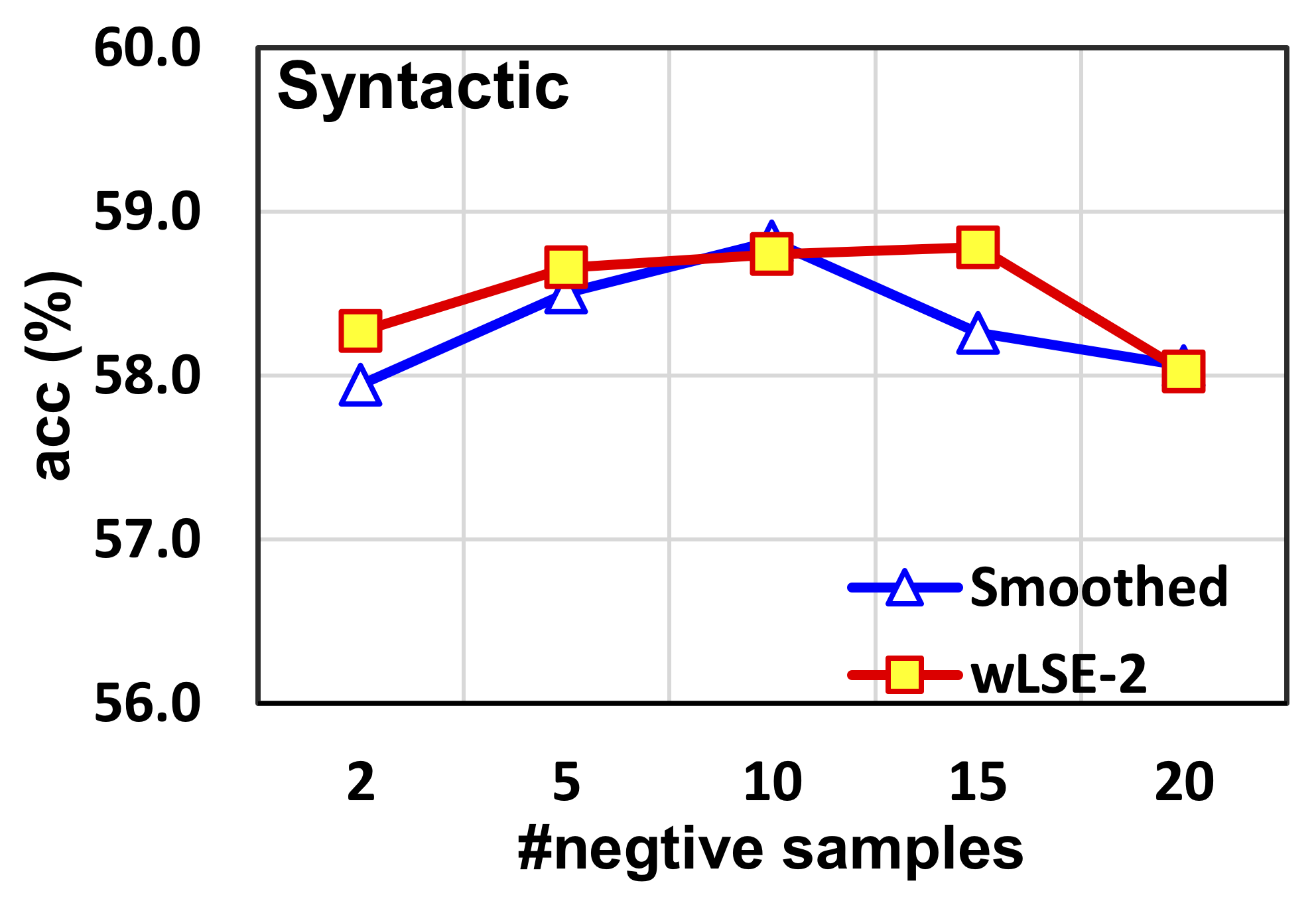}
\caption{SG}
\end{subfigure}
\hspace{0.6em}
\begin{subfigure}{0.3\textwidth} 
\includegraphics[width=\textwidth]{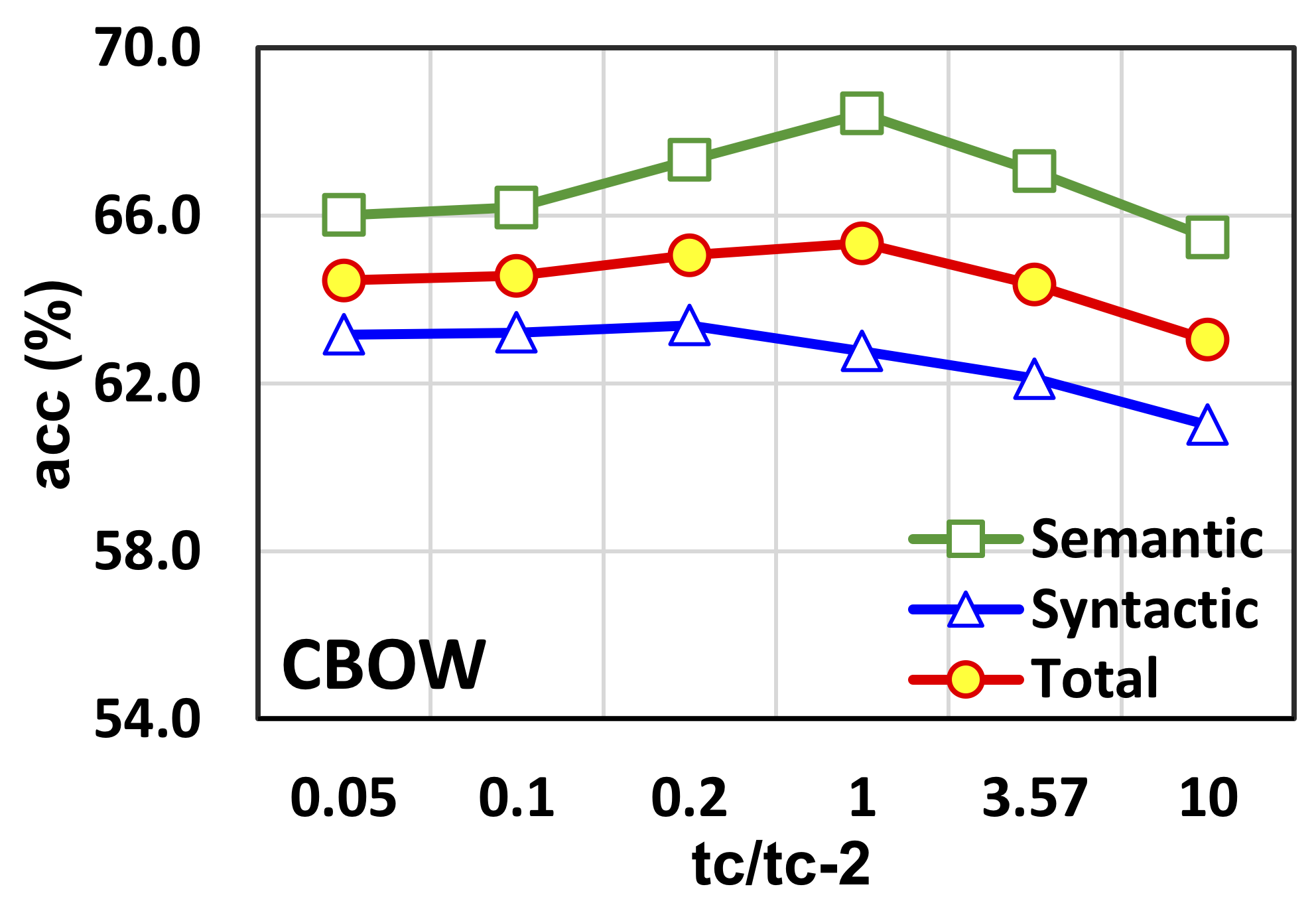} \\
\includegraphics[width=\textwidth]{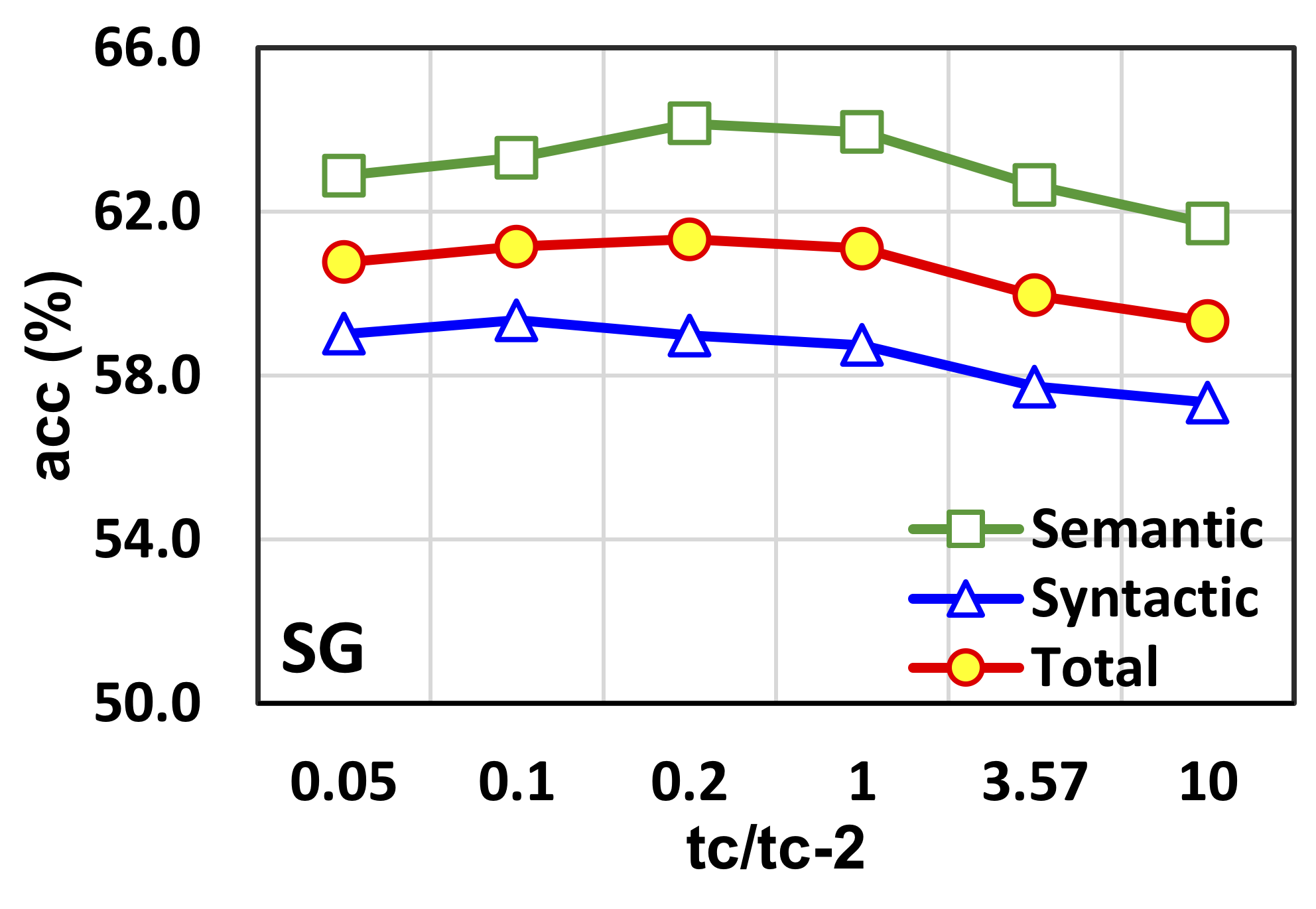}
\caption{Optimality}
\end{subfigure}
\caption{\label{neg_opt}Word analogy results (a) and (b) for number of negative samples and (c) for optimality. Smoothed and wLSE-2 represent $Uni^{3/4}$ and $Sub^{L2}$, $t_c$-2 means the sub-sampling rate of $Sub^{L2}$.}
\end{figure*}

\subsubsection{Task Description}
\label{sssec: Task3_Desp}

The task comes from the idea that arithmetic operations in a word vector space can be predicted: given three words $w_a$, $w_b$, and $w_c$, the goal is to find a word $w_d$ such that the relation $w_d:w_c$ is the same as the relation $w_b:w_a$. Semantic questions are in the form of ``Athens:Greece is as Berlin:German'' and syntactic ones are like ``dance:dancing is as fly:flying''. Here we choose the fourth word $\hat w_d$ by maximizing the cosine similarity such that $\hat w_d = \argmax_{w\in V} \,\cos\left( v_{w_b}-v_{w_a}+v_{w_c}, v_w\right)$~\citep{mikolov2013linguistic}.
We test the learned word vectors on the Google analogy dataset~\citep{mikolov2013distributed}, which contains 8,869 semantic questions and 10,675 syntactic ones.

\subsubsection{Results}
\label{sssec: Task3_results}

This task is our primary focus because it exposes interesting linear relationships between word vectors. Thus we conduct four sub-experiments to investigate four aspects of our noise distributions.

\textbf{Model Responses.} The two models SG and CBOW respond differently to our noise distributions as shown in Table\,\ref{Tresults}. When applying CBOW model on the three corpora, our noise distributions {\em Sub$^{L1}$} and {\em Sub$^{L2}$} can result in significant improvements compared with {\em Uni$^{3/4}$}, especially on semantic questions. Specifically, the accuracy of semantic questions is improved by 2 to 6 points, and for syntactic questions it is 1.5 to 2 points. As for the SG model, the improvements on semantic questions by {\em Sub$^{L1}$} and {\em Sub$^{L2}$} are still considerable (2 to 5 points). But on syntactic questions, {\em Uni$^{0.75}$} becomes competitive with {\em Sub$^{L1}$} and {\em Sub$^{L2}$} and is slightly better with BWLM and Wiki10 corpora. The reason may be that SG model is better at capturing semantic relationships between words compared with CBOW model. Still, it is safe to say that our noise distributions are better for SG in terms of the total accuracy.

\textbf{Number of Negative Samples.} Increasing the number of negative samples does not reduce the advantages of our noise distributions necessarily. We report the results of the task using various number of negative samples in Fig.\,\ref{neg_opt}\,(a) for CBOW and Fig.\,\ref{neg_opt}\,(b) for SG. Note that we only train the models on Wiki10 and compare {\em Sub$^{L2}$} with {\em Uni$^{3/4}$}.
For CBOW, {\em Sub$^{L2}$} outperforms {\em Uni$^{3/4}$} consistently with significant margins on both semantic and syntactic questions. For SG, though the two distributions are competitive with each other on syntactic questions, {\em Sub$^{L2}$} always performs better than {\em Uni$^{3/4}$} on semantic ones.

\textbf{Optimality.} Since our approach is built on assumptions and new concepts, we wonder whether the resulted $t_c$ is optimal. We select several values around $t_c$-2 and show the word analogy results in Fig.\,\ref{neg_opt}\,(c).
For CBOW, $t_c$-2 approaches the optimal point given the accuracy on semantic questions and the total dataset. For SG, the optimal point lies between $0.1\,t_c$-2 and $t_c$-2, with negligible advantages relative to {\em Sub$^{L2}$}. Notice that the point $3.57\,t_c$-2 corresponds to $10^{-5}$, showing much worse performance than {\em Sub$^{L2}$}. It indicates that trying a commonly used sub-sampling rate is inappropriate, and our approach is better.

\textbf{Scalability.} We apply our noise distributions in NCE, from which negative sampling originates, to train word vectors. The implementation comes from {\tt wang2vec}\footnote{\url{https://github.com/wlin12/wang2vec}} by \citet{ling2015two}, and we report the results of this task using CBOW.
We include the unigram distribution {\em Uni}~\citep{mnih2013learning} and the sub-sampled unigram distribution {\em Sub$^{1e-5}$} with a manually chosen threshold $10^{-5}$ for comparison. We draw three conclusions: (1) {\em Uni$^{3/4}$} indeed works much better than {\em Uni} as claimed in~\citep{mikolov2013distributed}; (2) {\em Sub$^{1e-5}$} results in considerable improvements compared with {\em Uni$^{3/4}$}, especially on semantic questions; (3) Our {\em Sub$^{L2}$} achieves the best performance consistently even with a larger vector size of 300. Note that even though {\em Sub$^{1e-5}$} or {\em Uni$^{3/4}$} performs better on syntactic questions with UMBC corpus, its results on semantic questions and the total dataset are much worse than {\em Sub$^{L2}$}. To this end, we believe that our approach is also scalable to the NCE related work.

\begin{table}[t!]
\centering
\captionsetup{belowskip=-1em}
\begin{threeparttable}
\begin{tabular}{|l|c|l|c|c|c|}
\hline
\bf Size & \bf Dim & \bf Noise & \bf Sem & \bf Syn & \bf Tot \\
\hline
\multirow{7}{*}{0.7\,B}
& \multirow{4}{*}{100}
& {\em Uni} & 36.2 & 47.5 &	42.5 \\
& & {\em Uni$^{3/4}$} & 44.8 & 50.5 & 47.9 \\
& & {\em Sub$^{1e-5}$} & 49.4 & 51.4 & 50.5 \\
& & {\em Sub$^{L2}$} & \bf 52.3 & \bf 51.8 & \bf 52.0 \\
\cline{2-6}
& \multirow{2}{*}{300}
& {\em Uni$^{3/4}$} & 46.4 & 58.3 & 53.0 \\
& & {\em Sub$^{L2}$} & \bf 55.0 & \bf 59.7 & \bf 57.6 \\
\hline
\multirow{7}{*}{1.0\,B}
& \multirow{4}{*}{100}
& {\em Uni} & 51.5 & 47.6 & 49.3 \\
& & {\em Uni$^{3/4}$} & 57.5 & 50.7 & 53.8 \\
& & {\em Sub$^{1e-5}$} & 61.9 & 51.1 & 56.0 \\
& & {\em Sub$^{L2}$} & \bf 63.5 & \bf 52.7 & \bf 57.6 \\
\cline{2-6}
& \multirow{2}{*}{300}
& {\em Uni$^{3/4}$} & 65.8 & 59.0 & 62.1 \\
& & {\em Sub$^{L2}$} & \bf 70.3 & \bf 60.8 & \bf 65.1 \\
\hline
\multirow{7}{*}{3.0\,B} 
& \multirow{4}{*}{100}
& {\em Uni} & 25.4 & 48.1 & 38.0 \\
& & {\em Uni$^{3/4}$} & 34.7 & 54.7 & 45.8 \\
& & {\em Sub$^{1e-5}$} & 37.1 & \bf 55.7 & 47.4 \\
& & {\em Sub$^{L2}$} & \bf 42.6 & 54.8 & \bf 49.4 \\
\cline{2-6}
& \multirow{2}{*}{300}
& {\em Uni$^{3/4}$} & 52.4 & \bf 62.3 & 57.9 \\
& & {\em Sub$^{L2}$} & \bf 62.0 & 61.8 & \bf 61.9 \\
\hline
\end{tabular}
\end{threeparttable}
\caption{\label{Tnce}The results of word analogy task using NCE for the training of word vectors. Each entry is the average result of 2 repeated experiments.}
\end{table}

\subsection{Extension of Semantics Quantification}
\label{ssec: ExtendSemQ}

\subsubsection{MSR Sentence Completion Task}
\label{sssec: MSRsc_Desp}

The task deals with incompletion sentences, {\em e.g.}, ``A few faint \underline{\qquad} were gleaming in a violet sky.'' with candidate answers ``tragedies, stars, rumours, noises, explanations'', and aims to choose a word ({\em e.g.}, ``stars'') to best complete the sentence. Several works evaluate word vectors on this task~\citep{mikolov2013efficient,mnih2013learning,liu2015learning} since it requires a combination of semantics and occasional logical reasoning. 
Most of them follow the same procedures of implementation described in~\citep{mnih2012fast}. Specifically, we can calculate the probabilities that a set of words $\mathcal{S}$ surrounding the blank to be the context of each candidate answer $w_{cd}$. Then the score of the candidate answer is the sum of these probabilities,
\begin{equation}
{\rm CM}(w_{cd}) = \frac{\sum_{w\in \mathcal{S}}\exp\left(v_{w}'^{\top}v_{w_{cd}}\right)}
{\sum_{w\in V}\exp\left(v_{w}'^{\top}v_{w_{cd}}\right)} ,
\end{equation}
and the highest score corresponds to the final answer for the question.

Since the conventional method ignores the syntactic structure of sentences, it should be biased to semantics. Thus, we modify the method with two steps: (1) applying sub-sampling on the words in the sentences (CM$^s$); and (2) using quantified semantics as weights to form a semantics weighted model (SWM) based on (1). Then we have
\begin{equation}
{\rm SWM}(w_{cd}) = \frac{\sum_{w\in \mathcal{S}}I_{sem}^w \exp\left(v_{w}'^{\top}v_{w_{cd}}\right)}
{\sum_{w\in V}\exp\left(v_{w}'^{\top}v_{w_{cd}}\right)} .
\end{equation}

\begin{table}[t!]
\centering
\captionsetup{belowskip=-1em}
\begin{threeparttable}
\begin{tabular}{|l|c|}
\hline
\bf Model & \bf Acc \\
\hline
LSA~\citep{zweig2012computational} & 49.0 \\
SG~\citep{mikolov2013distributed} & 48.0 \\
ivLBL~\citep{mnih2013learning} & 55.5 \\
SWE~\citep{liu2015learning} & 56.2 \\
\hline
\end{tabular} %
\begin{tabular}{|c|c|l|c|c|}
\hline
\bf Model & \bf Dim & \bf Score & $Uni^{3/4}$ & $Sub^{L2}$ \\
\hline
\multirow{6}{*}{\tt sg} 
 & \multirow{3}{*}{100}
 & CM & 49.0 & 49.4 \\
 & & CM$^s$ & 54.8 & 54.4 \\
 & & SWM & \bf56.0 & \bf56.5 \\
\cline{2-5}
 & \multirow{3}{*}{300}
 & CM & 49.9 & 49.5 \\
 & & CM$^s$ & 56.4 & 55.4 \\
 & & SWM & \bf58.0 & \bf57.9 \\
\hline
\multirow{6}{*}{\tt cbow} 
 & \multirow{3}{*}{100}
 & CM & 47.8 & 46.2 \\
 & & CM$^s$ & 55.3 & 54.3 \\
 & & SWM & \bf56.3 & \bf55.8 \\
\cline{2-5}
 & \multirow{3}{*}{300}
 & CM & 49.6 & 48.8 \\
 & & CM$^s$ & 56.4 & 56.1 \\
 & & SWM & \bf57.5 & \bf57.3 \\
\hline
\end{tabular}
\end{threeparttable}
\caption{\label{Tsentcom} The results of MSR sentence completion task by previous word representation models and our approach.}
\end{table}

\subsubsection{Results}
\label{sssec: MSRsc_results}

The setup of models is a little different: the size of context window for SG and CBOW is 10 and 5; the number of negative samples is 20 in both models; we train SG for 5 and 10 epochs when the size of word vectors is 100 and 300, while the number of epochs is 10 and 20 in CBOW; we use all the rest words in a sentence to form $\mathcal{S}$.

Our focus here is to popularize SWM rather than to compare the noise distributions. We show the results of this task by previous word presentation models and our approach in Table\,\ref{Tsentcom}. The bottom three previous models follow the conventional method. Accordingly, we draw two conclusions: (1) sub-sampling on the words in sentences results in significant improvements to the conventional method; and (2) SWM further improves CM$^s$ and beats previous word representation models with a vector size of 300, indicating the success of semantics quantification.




\section{Conclusions}
\label{sec: Conclusions}

We propose to employ a sub-sampled unigram distribution for better negative sampling, and design an approach to derive the required sub-sampling rate. Experimental results show that our noise distribution captures better linear relationships between words than the baselines. It adapts to different corpora and is scalable to NCE related work. The proposed semantics weighted model also achieves a success on the MSR sentence completion task. In summary, our work not only improves the quality of word vectors, but also sheds light on the understanding of Word2Vec.


\bibliography{emnlp2018}
\bibliographystyle{acl_natbib_nourl}

\end{document}